# Federated and Distributed Learning Applications for Electronic Health Records and Structured Medical Data: A Scoping Review

Siqi Li[1#], Pinyan Liu[1#], Gustavo G. Nascimento[2,3], Xinru Wang[1], Fabio Renato Manzolli Leite[2,3], Bibhas Chakraborty[1,4,5,6], Chuan Hong[6], Yilin Ning[1], Feng Xie[1,4], Zhen Ling Teo[7], Daniel Shu Wei Ting[1,7], Hamed Haddadi[8], Marcus Eng Hock Ong[4,9], Marco Aurélio Peres[2,3,4^], Nan Liu[1,4,10^]*

[1] Centre for Quantitative Medicine, Duke-NUS Medical School, Singapore, Singapore

[2] National Dental Research Institute Singapore, National Dental Centre Singapore, Singapore

[3] Oral Health Academic Clinical Programme, Duke-NUS Medical School, Singapore, Singapore

[4] Programme in Health Services and Systems Research, Duke-NUS Medical School, Singapore, Singapore

[5] Department of Statistics and Data Science, National University of Singapore, Singapore, Singapore

[6] Department of Biostatistics and Bioinformatics, Duke University, Durham, NC, USA

[7] Singapore National Eye Centre, Singapore, Singapore Eye Research Institute, Singapore

[8] Department of Computing, Imperial College London, London, England, United Kingdom

[9] Department of Emergency Medicine, Singapore General Hospital, Singapore, Singapore

[10] Institute of Data Science, National University of Singapore, Singapore, Singapore

[#]Joint first author

[^]Joint senior author

* Correspondence: Nan Liu, Centre for Quantitative Medicine, Duke-NUS Medical School, 8 College Road, Singapore 169857, Singapore. Phone: +65 6601 6503. Email: liu.nan@duke-nus.edu.sg

## Abstract

### Objectives

Federated learning (FL) has gained popularity in clinical research in recent years to facilitate privacy-preserving collaboration. Structured data, one of the most prevalent forms of clinical data, has experienced significant growth in volume concurrently, notably with the widespread adoption of electronic health records in clinical practice. This review examines FL applications on structured medical data, identifies contemporary limitations and discusses potential innovations.

### Materials and Methods

We searched five databases, SCOPUS, MEDLINE, Web of Science, Embase, and CINAHL, to identify articles that applied FL to structured medical data and reported results following the PRISMA guidelines. Each selected publication was evaluated from three primary perspectives, including data quality, modeling strategies, and FL frameworks.

### Results

Out of the 1160 papers screened, 34 met the inclusion criteria, with each article consisting of one or more studies that used FL to handle structured clinical/medical data. Of these, 24 utilized data acquired from electronic health records, with clinical predictions and association studies being the most common clinical research tasks that FL was applied to. Only one article exclusively explored the vertical FL setting, while the remaining 33 explored the horizontal FL setting, with only 14 discussing comparisons between single-site (local) and FL (global) analysis.

**Conclusions**

The existing FL applications on structured medical data lack sufficient evaluations of clinically meaningful benefits, particularly when compared to single-site analyses. Therefore, it is crucial for future FL applications to prioritize clinical motivations and develop designs and methodologies that can effectively support and aid clinical practice and research.

## 1. Introduction

The digitization of electronic health records (EHRs) has enabled the analysis of data from multiple centers, which allows for comparisons across populations and settings or data combinations to improve statistical power and generalizability[1]. Data sharing, a traditional strategy for forming cross-regional partnerships in the healthcare industry, has been shown to benefit the reproducibility of research, cost-efficiency, prevention of redundancies, and accelerations of discovery and innovation[2]. However, such cooperation raises concerns about data privacy, making privacy-preserving techniques an area of interest in this field.

Federated learning (FL) has gained popularity in healthcare as a technique for maintaining privacy[3]. FL is a machine learning setting where multiple entities (clients) collaborate in solving a modeling problem, with the data for each client stored locally and not exchanged or transferred[4]. Prior to the introduction of the term FL[5], statisticians had been researching privacy-preserving statistical algorithms using terms such as "distributed learning"[6,7] or "distributed algorithms"[8,9]. Despite differences in terminology, these algorithms all contribute to reducing barriers posed by privacy regulations across different countries and regions, enabling researchers to collaborate more effectively and efficiently.

Although there exist a number of FL applications in the medical field, most of these early adoptions have focused on unstructured data, particularly image data[10]. Structured data, which constitute a significant part of clinical data with the advent of large-scale EHRs, has been relatively underexplored in FL settings. Structured data-based FL differs from image data-based

FL in several ways, including sample sizes, data structures, modeling methodologies, research topics, and study designs.

Although several reviews have been conducted on FL applications in healthcare, they often discuss clinical data in general terms without delving into specific data types[6,10,11]. Furthermore, while some of these reviews address technical issues in-depth, they fall short in presenting their conclusions from the perspective of clinical applications. In response, we conducted a scoping review to summarize and examine current FL applications on structured clinical data, emphasizing the advantages of FL for clinical research. Our aim is to provide insights and suggestions for future FL applications in clinical decision-making.

## 2. Materials and Methods

### 2.1 Search strategy and selection criteria

We conducted a review following the 2020 PRISMA[12] guidelines for reporting systematic reviews and searched for published articles that used FL frameworks to solve clinical/biomedical questions using structured data. We searched the SCOPUS, MEDLINE, Web of Science, Embase, and CINAHL databases using a combination of search terms, including "electronic health records", "EHR", "electronic medical records", "EMR", "registry/registries", "tabular", "federated learning", "distributed learning" and "distributed algorithms". A detailed search strategy is presented in eTable 1 of the Supplementary.

The final search was conducted on Aug 23rd, 2022. After removing duplicates, two reviewers (S.L. and P.L.) independently screened articles based on their titles and abstracts, with a third reviewer (F.X.) resolving any conflicts. The publications selected in the first round of screening underwent full-text examination to ensure they met the exclusion criteria: not using structured data, not using federated learning, not being a research article, not having full text available, and not addressing biomedical/clinical research data/questions.

### 2.2 Data extraction

We extracted information from the selected publications from three perspectives: data (cohort descriptive analysis, outcome, sample size per site and total, number of participating sites, data types, data public availability and number of features), modeling (task and goal, modeling approach, hyperparameter methods, model performance metrics), and FL frameworks (unit of federation, participating countries/regions, FL structure, FL topology, one-shot or not, evaluation metrics, convergence analysis, solution for heterogeneity, FL and local model comparison and code availability).

## 3. Results

Our search strategy yielded 1160 articles, and after removing 601 duplicate records, 559 articles were screened based on title and abstract. Sixty-two out of 559 articles were left for full-text screening, and 34 articles were included for this review. One article may present more than one studies, depending on datasets, models, and FL frameworks applied. For instance, Halim et al.[13] carried out two studies using the same dataset but different outcomes (one binary and one multi-label); Sadilek et al.[14] undertook seven studies with seven different datasets. Consequently, as shown and summarized in Table 1, the total number of studies for all 34 included papers is 72, and full details can be found in Supplementary eTable 2.

### 3.1 General characteristics of included papers

Twenty-nine of the 34 papers included in this review were published in or after 2020, indicating a recent surge of interest in FL for clinical research with structured data. The total sample size of the FL settings in each study varied from 141 ([15]) to 4,408,710 ([16]), and the number of clients (participating sites) ranged from 2 ([17–20]) to 314 ([11]). There are two types of participating clients: those who used artificially partitioned datasets and those who used real isolated datasets. Only one article ([21]) assumed a vertical FL setting, in which all datasets share the same sample space but differ in features[22], while the others discussed horizontal FL, meaning that datasets share the same feature space but have different samples[22].

### 3.2 Dataset characteristics

Twenty-five out of 34 papers used structured data derived from EHRs, primarily containing patient demographics, vital signs, and other features commonly found in hospital records. Five articles ([18,20,23–25]) studied clinical cohorts that underwent long-term interventions or follow-up. Only 38.9% (28/72) of the studies provided descriptive analyses of their data, and 62.5% (45/72) used publicly accessible datasets. Five studies used datasets with more than 1000 features, while the remaining studies mostly (58.3% = 42/72) used data with fewer than 41 features.

### 3.3 Modeling characteristics

In this review, we classified a study as ‘prediction’ if its primary goal was to predict an outcome and report performance metrics, and a study as ‘association modeling’ if its primary goal was to investigate the relationship between covariates and outcome(s) by reporting estimates of coefficients or odds ratios. These two types of studies differ in statistical inference, with the latter focusing on uncertainty measurements while the former does not. Another type of task is phenotyping, which is unsupervised and aimed at deriving research-grade phenotypes from clinical data[26]. Of all 72 studies, 40 (55.6%) performed prediction tasks, 20 (27.8%) investigated associations, and 9 (12.5%) conducted phenotyping. A majority of the studies (69.4% = 50/72) investigated binary outcomes, and logistic regression was the most frequently used model (25.0% = 18/72). Various types of neural networks were used in 26.4% (19/72) of the studies.

### 3.4 FL architectures

We categorized FL algorithms into one-shot and non-one-shot based on whether a FL algorithm required one or multiple rounds of communication (intermediate parameters needed to be transferred once or more times among participants). Our observations showed that 26.4% (19/72) of the studies used one-shot FL and 70.8% (51/72) used non-one-shot FL, with 74.5% (38/51) using a centralized FL topology and 25.5% (13/51) using a decentralized FL topology. We summarized the frameworks used in all studies and created a plot (Figure 2) to illustrate their applications and interdependencies. Of all studies, 27.8% (20/72) used FL frameworks that adopted one or more particular solutions to handle data heterogeneity.

### 3.5 FL performance evaluation

Based on the articles reviewed for this study, FL frameworks were frequently evaluated for their computational performance, such as rounds of communications, number of iterations, and computation time, but less frequently on their performance compared to conventional single-site analyses. Only 41.2% of (14/34) the papers specifically discussed comparisons between local and FL models, and among these, 11 reported that FL models had some advantages over local models.

## 4. Discussion

The use of FL for structured medical data has been extensively explored, but certain aspects of such studies require special attention. Our review indicates that FL is sometimes applied without first assessing its actual benefits. A critical consideration is whether FL can advance medical research objectives that are not achievable through single-site analyses. In this section, we provide a detailed discussion on when healthcare researchers should consider FL, additional precautions that should be taken, and technical details related specifically to clinical structured data.

### 4.1 Is FL necessary?

Before diving into technical details, we examine the necessity of FL for conducting research on structured medical data. We will start with discussing why and how FL is anticipated to yield additional meaningful results beyond those produced by pre-existing local models. We will then discuss how data can affect the feasibility and efficacy of FL.

Cross-silo and cross-device are two distinct FL settings that can significantly impact the design of FL algorithm[27]. Cross-silo design aims to facilitate collaboration among various organizations, such as hospitals and institutions, while cross-device design enables collaboration among large populations of mobile devices[27]. When designing and implementing cross-silo FL, it is important to consider the presence of pre-existing local models, in addition to the technical distinctions highlighted in Wang et al.[27]. This is particularly crucial given that the vast amount of information available in today's large-scale EHRs.

The comparison of the performance of FL models to local models has therefore become a natural point of discussion. Some papers that benchmark or develop FL frameworks in a broader context have highlighted the relevance and importance of such comparisons[28,29]. However, in this review, such comparisons were only observed in 14 out of 34 papers. For example, Cui et al.[30] developed a new FL framework called FeARH and evaluated its performance using artificially partitioned data. Although they provided a performance comparison of the central, FL baseline, and FeARH models, future researchers may find such demonstrations insufficient, as obtaining a ground truth central model is often impossible in real-world practice.

Unlike unstructured data such as images, structured clinical data often contains features that can vary at the definition stage owing to differences in clinical practice across institutions[31,32]. For example, diabetes is characterized by elevated levels of glucose in the bloodstream, and it can be diagnosed by fasting or random plasma glucose, each of which has a different cut-off point. Such inconsistency in disease diagnosis may result in hidden heterogeneity that is undetectable by downstream statistical analysis. Therefore, proper preparation and data harmonization across participating sites are required prior to implementing FL.

Based on the findings of this review, we recommend that two broad goals should be achieved for FL applications with structured healthcare data. First, FL models should outperform locally developed models for at least one participating site. This can be demonstrated by improved accuracy in prediction tasks, narrower confidence intervals in effect size estimations, or the discovery of new phenotypes in phenotyping tasks. For instance, Dayan et al.[33] showed that the best FL model for predicting 24-hour oxygen treatment for COVID-19 patients outperformed all 20 local models trained independently by each site. In another example, Kim et al.[19] proposed a federated tensor factorization framework that successfully identified new phenotypes that were not captured in any of the local phenotyping analyses[19]. Second, for prediction tasks in particular, FL models should exhibit better stability and generalizability than at least one local model, as demonstrated by achieving lower performance variation than some local models. Dayan et al.[33] demonstrated that the best FL model exhibited the highest level of generalizability among all local models, as measured by average area under the curve (AUC).

### 4.2 Technical details and challenges

#### FL algorithms: statistical versus engineering approaches

Both the statistics and engineering communities have conducted research on FL, but a significant distinction between them is the property of model agnosticism. Statistics-based FL algorithms usually involve model-specific statistical modeling, meaning that a single algorithm is usually applied for one type of model. For instance, as illustrated in Figure 2, a federated Cox regression might only be conducted using statistics-based FL methods such as ODAC[23], ODACH[25], and SurvMaximin[1]. By contrast, most engineering-based approaches have been developed in a model-agnostic manner, allowing for the use of a single FL framework for different machine learning models by employing the appropriate loss functions. For instance, in Sadilek et al.[14], FedAvg[5] has been applied to neural networks, logistic regression and generalized linear models for binomial responses with log link.

To the best of our knowledge, there are no direct comparisons in current literature between statistics-based and engineering-based FL algorithms, and their precise advantages and disadvantages have not been adequately assessed. One apparent advantage of engineering-based methods is their model-agnostic property, which allows these algorithms to be directly applied to a wide range of commonly used models without the need for additional designs. Moreover, engineering-based solutions may be more resilient to model misspecification as most existing statistics-based FL algorithms focus on linear relationships that may not hold for real-world data.

Although statistics-based FL algorithms require one-to-one development and validation, they have an advantage over engineering-based FL in uncertainty measurements. In FL studies that aim to estimate parameters of interest such as the association between exposure and outcome, it is desirable to report the associated confidence intervals. Statistics-based FL methods have an advantage in estimating uncertainty measures in a distributed manner when the asymptotic distribution of an estimator is available and a closed-form formula for the variance of estimated parameters exists[23,25,34,35]. In the absence of these conditions, bootstrap can serve as a flexible alternative to estimate standard errors for both statistics- and engineering-based FL. However, it is important to note that a simple bootstrap strategy may produce inconsistent results if the estimator is non-smooth, requiring additional measures to ensure effectiveness[36]. While few existing FL studies have examined the uncertainty of estimated associations, future investigations may consider these potential limitations.

## Statistical heterogeneity: challenges and potential benefits

In many cases, FL frameworks assume that data is independently and identically distributed (i.i.d.)[3], and may struggle in heterogeneous settings[37,38]. However, opinions also hold that these constraints may not have as much of an impact as anticipated[3], and such FL algorithms can still achieve good results with non-i.i.d. data[3]. For example, Vaid et al.[39] reported that FedAvg outperformed local models even though the patient demographics were heterogenous. Therefore, we recommend that future researchers benchmark classic i.i.d.-based FL frameworks with heterogeneous data to provide empirical evidence of their effectiveness. It is also noteworthy that while heterogeneity is generally considered a challenge for supervised learning models, there is evidence to suggest that it can be beneficial for certain unsupervised tasks[40].

## Convergence analysis for optimization

Almost all FL algorithms aim to estimate the parameters of interest by solving an optimization problem, and convergence is crucial to ensure accurate estimation. If an FL algorithm frequently fails to converge, it may not be suitable for real-world data. However, convergence analysis has not been adequately addressed in the current FL literature of structured healthcare data. Of the 19 papers in this review proposing new FL frameworks, only five ([19,41–44]) discussed framework convergence. Stochastic gradient descent (SGD) is a popular choice for smooth optimization when the assumptions such as the existence of lower bounds, Lipschitz smoothness, and bounded variance are met[45]. However, SGD-based FL algorithms no longer function for non-smooth cases, requiring unique strategies as developed in some studies[43,46].

**Data privacy: differential privacy and blockchain**

Although FL enables model training without exchanging or sharing data, adversaries can still analyze the differences in related parameters trained and uploaded during the FL process to obtain private information[47]. To address this data leakage issue, differential privacy (DP) techniques[48] have been integrated with FL frameworks to add artificially controlled noise[11] before, during, or after model training[49]. DP has also been combined with generative adversarial network (GAN) to generate synthetic data that can be shared with collaborators. This approach is particularly useful for structured data[50], since GAN's have a greater flexibility in modeling distributions compared to their statistical counterparts[51].

While only five articles ([14,33,42,50,52]) in this review addressed the integration of DP with FL, broader reviews of FL in computer science[49,53] have shown that it is a popular topic for general FL applications. As FL continues to be adopted for large-scale and widespread medical applications, the integration of FL with technologies such as DP is likely to become increasingly important in the future.

### 4.3 Calling for standardized and open-source pipelines

Our review revealed that many existing FL studies do not provide adequate information on their model selection and data preprocessing procedures. The absence of open-source codes also hinders the reproducibility and future applications of FL. To address these issues, we recommend that future FL research in healthcare involve close collaboration with healthcare professionals to establish more standardized processes and assessments.

### 4.4 Limitations

The goal of this review is to provide a comprehensive overview of FL applications on structured clinical data, and thus we did not delve into detailed analysis of the mathematical and technological aspects of FL frameworks, particularly with respect to optimization.

## 5. Conclusion

FL applications on structured medical data are still in their early stages, with most studies primarily focused on prediction tasks and lacking robust demonstrations of clinically significant results. Further investigation into a combination of engineering- and statistics-based FL algorithms may open up novel possibilities. Additionally, our review highlights the significance of establishing standardized methodologies and protocols, as well as releasing open-source codes, to promote reproducibility and transparency in future FL research in healthcare.

**Figure 1** Preferred Reporting Items for Systematic Reviews (PRISMA) flow diagram

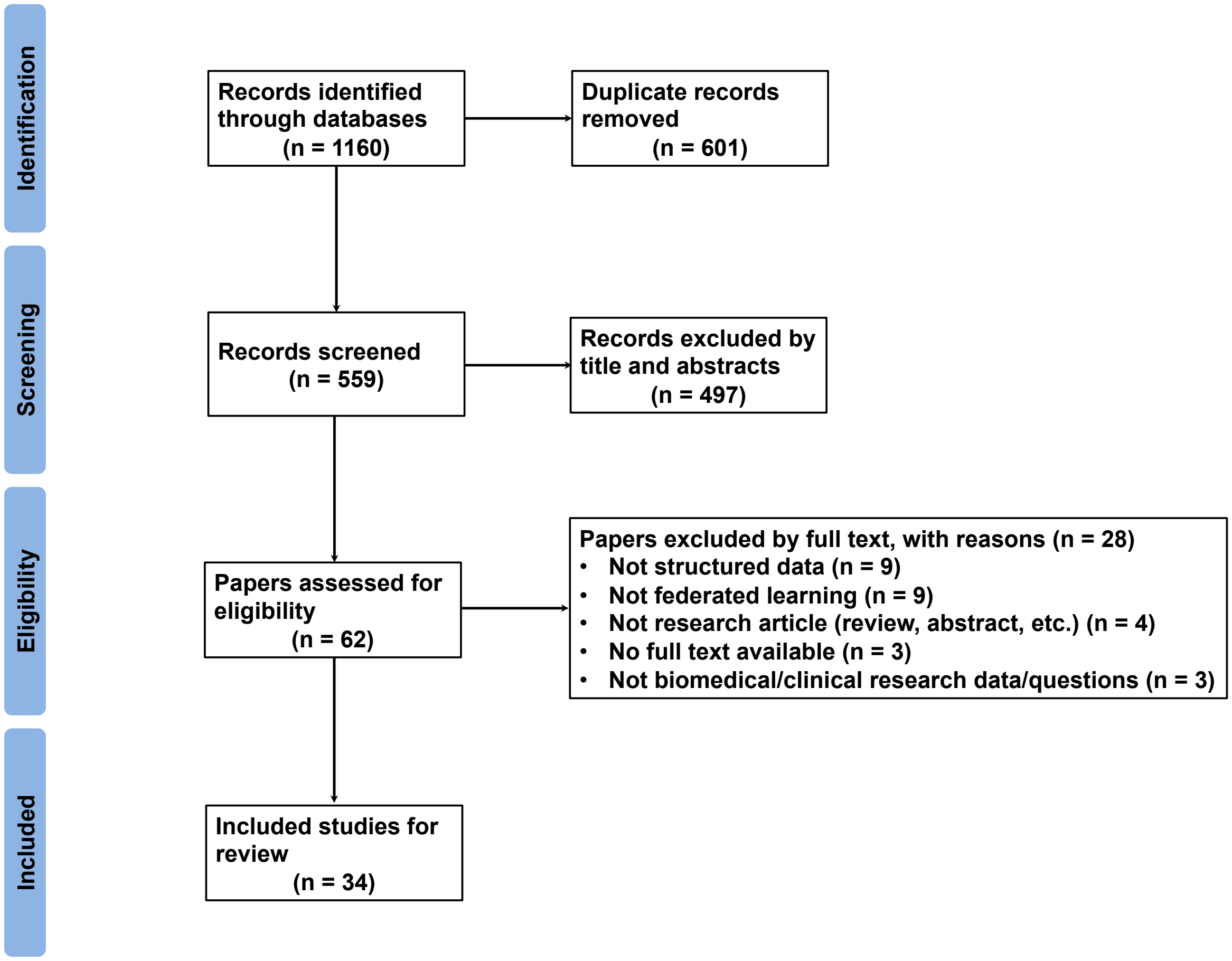

**Figure 2** Visualization of FL frameworks utilized in the 34 papers included in this review, classified into two categories: statistics-based and engineering-based FL as discussed in 4.3.

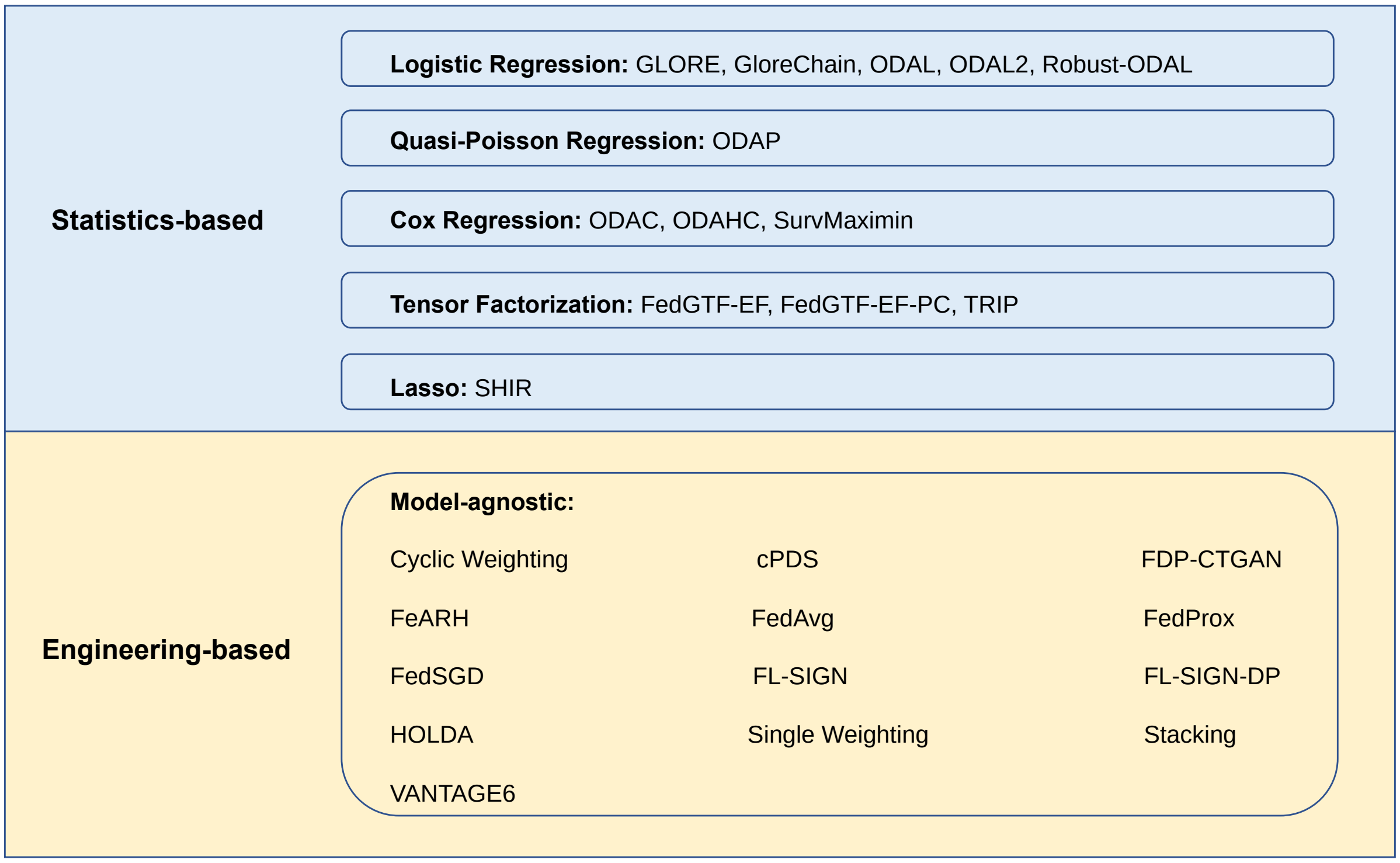

**Table 1** Summary of information extraction table.

| Data Characteristics | No. of Studies (out of 72) | Examples |
|---|---|---|
| Provide cohort descriptive analysis | 28 (38.9%) | A summary statistics table describing patient population was provided in [54] |
| Patient cohort size (total) | | |
| <= 50,000 | 53 (73.6%) | The total patient cohort size used in [33] was 16,148 |
| 50,000 - 100,000 | 3 (4.2%) | The total patient cohort size used (in the first study) in [54] was 70,818 |
| > 100,000 | 11 (15.3%) | The total patient cohort size used in [55] was 257,571 |
| Not available | 5 (6.9%) | No such details provided in [56] |
| No. clients | | |
| <= 5 | 35 (48.6%) | The number of clients in [30] was 8 |
| 6 - 35 | 20 (27.8%) | The number of clients in [44] was 20 |
| > 35 | 9 (12.5%) | The number of clients in [42] was 314 |
| Cross-device | 4 (5.6%) | The 1st, 2nd, 5th and 6th studies in [14] used cross-device (cross-patient) setting |
| Not available | 4 (5.6%) | No such details provided in [50] |
| No. features | | |
| <= 40 | 42 (58.3%) | The number of features in [39] was 23 |
| 41-1000 | 10 (13.9%) | The number of features in [16] was 85 |
| > 1000 | 5 (6.9%) | The number of features in [30] was 2931 |
| Not available | 15 (20.8%) | No such details provided in [57] |

| | | | |
|---|---|---|---|
| Outcome | | | |
| | Binary | 50 (69.4%) | In-hospital mortality was used as outcome in [42] |
| | Survival | 5 (6.9%) | Cancer survival time was used as outcome in [18] |
| | Other | 16 (22.2%) | Frequency of serious adverse events (count) was used as outcome in [54] |
| | Not available | 1 (1.4%) | No such details provided in [50] |
| Data public availability | | | |
| | Yes | 45 (62.5%) | eICU data was used in [58] |
| | No | 27 (37.5%) | Data used in [8] is not publicly available |

| **Model Characteristics** | | **No. of Studies (out of 72)** | **Examples** |
|---|---|---|---|
| Task | | | |
| | Prediction | 40 (55.6%) | Prediction of COVID mortality risk[1] |
| | Association study | 20 (27.8%) | Association between length of stay in COVID-19 patients with various patient characteristics[35] |
| | Phenotyping | 9 (12.5%) | Extraction of meaningful medical concepts[43] |
| | Model misconduct detection | 3 (4.2%) | A generalizable approach to identify model misconducts in FL[15] |
| Modeling approach | | | |
| | Logistic regression | 18 (25.0%) | Proposed and evaluated a one-shot distributed algorithm for logistic regression for heterogenous data[55] |
| | Cox regression | 6 (8.3%) | Proposed and evaluated a one-shot distributed algorithm for Cox regression[23] |
| | Neural networks (NN) | | |

| | | |
|---|---|---|
| 1) General | 10 (13.9%) | 3-layer fully connected NN[30] |
| 2) Autoencoder | 4 (5.6%) | 5-layer fully connected denoising autoencoder[58] |
| 3) Perceptron | 3 (4.2%) | 3-layer MLP[39] |
| 4) Deep learning | 2 (2.8%) | TabNet [52] |
| SVM | 4 (5.6%) | Proposed and evaluated a FL framework for soft-margin $l_1 -$ regularized sparse SVM[46] |
| Tensor factorization | 9 (12.5%) | Proposed and evaluated a federated tensor factorization method[19] |
| Other | 15 (20.8%) | XGBoost[20] |
| Not available | 1 (1.4%) | No such details provided in [50] |

| **FL Characteristics** | | **No. of Studies (out of 72)** | **Examples** |
|---|---|---|---|
| Unit of federation | | | |
| | Cross silo | 64 (88.9%) | Three healthcare facilities within the OneFlorida Clinical Research Consortium[35] |
| | Cross patient (device) | 4 (5.6%) | Four studies reported cross patient results[14] |
| | Both | 3 (4.2%) | Three studies reported both cross patient and cross silo results[14] |
| | Not available | 1 (1.4%) | No such details provided in [50] |
| Participants | | | |
| | Real isolated sites | 37 (51.4%) | Data from 20 institutes across the globe[33] |
| | Artificial partitions | 32 (44.4%) | Ten sites formed by random splitting one dataset[36] |
| | Not available | 3 (4.2%) | No such details provided in [44] |
| FL Topology | | | |

| | | |
|---|---|---|
| Centralized | 38 (52.8%) | FL-SIGN and FL-SIGN-DP[42] |
| Decentralized | 13 (18.1%) | Decentralized stochastic gradient descent (DSGD) and tracking (DSGT)[44] |
| One-shot | 19 (26.4%) | ODAL[8] |
| Not available | 2 (2.8%) | No such details provided in [21] |
| Solution(s) for heterogenous data | | |
| Yes | 20 (27.8%) | Employed HinSAGE which introduces extra weight matrices for heterogeneous graph[13] |
| No | 52 (72.2%) | Not available |
| Local vs. FL comparison | | |
| Yes | 37 (51.4%) | Compared performances of local and FL models, where FL model outperformed local models[33] |
| No | 33 (45.8%) | Not available |
| Not applicable | 2 (2.8%) | Vertical instead of horizontal FL was used in[21] |

## Supplementary

**eTable 1**. Search Strategies

| Database | Search Strategy |
|---|---|
| PubMed | ("EHR"[Title/Abstract] OR "electronic health record*"[Title/Abstract] OR "EMR"[Title/Abstract] OR "electronic medical record*"[Title/Abstract] OR registry[Title/Abstract] OR registries[Title/Abstract] OR tabular[Title/Abstract]) AND ("federated learning"[Title/Abstract] OR "distributed learning"[Title/Abstract] OR "distributed algorithm*"[Title/Abstract] OR "federated"[Title/Abstract]) |
| Scopus | ( TITLE-ABS-KEY ( ehr ) OR TITLE-ABS-KEY ( "electronic health record*" ) OR TITLE-ABS-KEY ( emr ) OR TITLE-ABS-KEY ( "electronic medical record*" ) OR TITLE-ABS-KEY ( registry ) OR TITLE-ABS-KEY ( registries ) OR TITLE-ABS-KEY ( tabular )) AND ( TITLE-ABS-KEY ( "federated learning" ) OR TITLE-ABS-KEY ( "distributed learning" ) OR TITLE-ABS-KEY ( "distributed algorithm*" ) OR TITLE-ABS-KEY ( federated ) ) |
| Web of Science | (AB = (EHR OR "electronic health record*" OR EMR OR "electronic medical record*" OR registry OR registries OR tabular) OR TI = (EHR OR "electronic health record*" OR EMR OR "electronic medical record*" OR registry OR registries OR tabular) OR KP = (EHR OR "electronic health record*" OR EMR OR "electronic medical record*" OR registry OR registries OR tabular)) AND ( AB = ("federated learning" OR "distributed learning" OR "distributed algorithm*" OR "federated") OR TI = ("federated learning" OR "distributed learning" OR "distributed algorithm*" OR "federated") OR KP = ("federated learning" OR "distributed learning" OR "distributed algorithm*" OR "federated")) |
| Embase | ('ehr':ti,ab,kw OR 'electronic health record*':ti,ab,kw OR 'emr':ti,ab,kw OR 'electronic medical record*':ti,ab,kw OR registry:ti,ab,kw OR registries:ti,ab,kw OR tabular:ti,ab,kw) AND ('federated learning':ti,ab,kw OR 'distributed learning':ti,ab,kw OR 'distributed algorithm*':ti,ab,kw OR 'federated':ti,ab,kw) |
| CINAHL | Use ("EHR" OR "electronic health record*" OR "EMR" OR "electronic medical record*" OR registry OR registries OR tabular) AND ("federated learning" OR "distributed learning" OR "distributed algorithm*" OR "federated") for both Title OR Abstract |

**eTable 2**. Full information extraction table

(a) Data details

| Citation | No. studies | Descriptive analysis | Outcome | Patient cohort size (each site) | Patient cohort size (total) | No. sites | Types of structured data | Data public availability | No. features |
|---|---|---|---|---|---|---|---|---|---|
| Edmondson et al., 2022[35] | 2 | Yes | Length of stay (days) of COVID-19 patients (count) | 264 - 550 | 1049 | 3 | demographics, diagnosis, medical history | OneFlorida https://www.pcori.org/research-results/2015/oneflorida-clinical-research-consortium | 15 |
| | | Yes | Length of stay (days) of COVID-19 patients (count) | NA - 5029 | 45,169 | 38 | demographics, diagnosis, medical history | No | 15 |
| Wang et al., 2022[1] | 1 | No | COVID-19 mortality risk (binary) | NA | 83,178 | 17 | demographics, diagnosis, lab results | 4CE https://covidclinical.net/data/index.html | 19 |
| Kuo et al., 2022[15] | 3 | No | Presence of disease (binary) | each 25% of the total | 1253 | 4 | demographics, diagnosis, symptoms | Edinburg Myocardial Infarction https://doi.org/10.1093/oxfordjournals.eurheartj.a015035 | 9 |
| | | No | Presence of cancer (binary) | each 25% of the total | 141 | 4 | NA | Cancer Biomarker (“CA”) https://www.routledge.com/Statistical-Evaluation-of-Diagnostic-Performance-Topics-in-ROC-Analysis/Zou-Liu-Bandos-Ohno-Machado-Rockette/p/book/9781439812228 | 2 |
| | | No | Presence of infection (binary) | each 25% of the total | 157,493 | 4 | NA | No | 25 |
| | | No | The case of diabetes kidney disease (binary) | In-network sites: 1974 - 9386 | 17,455 | 15 | Clinical Classification Software (CSS) coded feature (grouping of ICD codes) | Cerner Health Facts EMR Data https://uthsc.edu/cbmi/data/cerner.php | 283 |
| Islam et al., 2022[59] | 2 | No | The case of diabetes kidney disease (binary) | Out-of-network siloed sites: 239 - 1900 | 17,455 | 16 | Clinical Classification Software (CSS) coded feature (grouping of ICD codes) | Cerner Health Facts EMR Data https://uthsc.edu/cbmi/data/cerner.php | 283 |
| Edmondson et al., 2021[54] | 2 | Yes | CHOP: total number of hospitalizations (count) | 5,456 - 13,074 | 70,818 | 6 | demographics | No | 6 |
| | | Yes | FOLFIRI: frequency of serious adverse events (SAEs) (count) | 48 - 386 | 660 | 3 | demographics, Charlson comorbidity index (CCI) | OneFlorida https://www.pcori.org/research-results/2015/oneflorida-clinical-research-consortium | 6 |
| Dayan et al., 2021[33] | 1 | Yes | COVID-19 patient oxygen therapy after ED admissions (binary) | 24 - 2,994 | 16,148 | 20 | demographics, vital signs, lab results, chest X-ray (CXR) images | Data for training at local sites are not public. Data from the independent validation sites are maintained by CAMCA, and access can be requested by contacting Q.L. | 20 (1 image 224 x224) |
| Kim et al., 2021[60] | 1 | No | In hospital mortality (binary) | NA | 3,146 | 8 | demographics | No | 4 |
| Cui et al., 2021[30] | 1 | No | Patient mortality (binary) | NA | 30,760 | 8 | medication prescriptions | eICU Collaborative Research Database https://eicu-crd.mit.edu/ | 2913 |
| Rajendran et al., 2021[17] | 2 | Yes | Presence of lung cancer or COPD (binary) | 6301 - 16,836 | 23,137 | 2 | demographics, exposures | No | 7 |
| | | Yes | Presence of lung cancer or COPD (binary) | 6301 - 16,836 | 23,137 | 2 | demographics, exposures | No | 7 |
| Geleijnse et al., 2020[18] | 1 | Yes | Cancer survival time (survival) | 7766 - 33,864 | 41,633 | 2 | demographics, diagnosis, treatments | The Taiwan Cancer Registry (TCR) https://ghdx.healthdata.org/record/taiwan-cancer-registry-incidence-and-mortality-1980-2007 The Netherlands Cancer Registry (NCR) https://iknl.nl/en/ncr-data | 8 |
| | | Yes | Time to the first occurrence of AMI (survival) | 59,861 - 69,164 | 255,595 | 4 | demographic, vital signs, diagnosis | CCAE https://cashe.psu.edu/resources/marketscan/ MDCD, MDCR https://www.ibm.com/products/marketscan-research-databases/databases Optum https://ssrc.indiana.edu/data/optum.html | 8 |
| Duan et al., 2020[23] | 2 | Yes | Time to the first occurrence of stroke (survival) | 59,861 - 69,164 | 255,595 | 4 | demographic, vital signs, diagnosis | CCAE https://cashe.psu.edu/resources/marketscan/ MDCD, MDCR https://www.ibm.com/products/marketscan-research-databases/databases Optum https://ssrc.indiana.edu/data/optum.html | 8 |
| Duan et al., 2020[34] | 1 | Yes | Risk of fetal loss (binary) | 3081 | 30,810 | 10 | demographics, medication prescriptions | UPHS (with request) https://uphsnet.uphs.upenn.edu/DACReportRequest/ | 5 |
| Tong et al., 2020[55] | 1 | Yes | The occurrence of the respective diagnosis codes (binary) | 1976 - 69,164 | 257,571 | 5 | diagnosis | CCAE https://cashe.psu.edu/resources/marketscan/ MDCD, MDCR https://www.ibm.com/products/marketscan-research-databases/databases Optum https://ssrc.indiana.edu/data/optum.html | 6 |
| | | | | | | | | JMDC https://www.eng.phm-jmdc.com | |
| Duan et al., 2021[24] | 1 | Yes | Time to the first diagnosis of ADRD (survival) | 983 - 12,181 | 16,456 | 3 | demographics, vital signs, diagnoses, and medication prescriptions | OneFlorida https://www.pcori.org/research-results/2015/oneflorida-clinical-research-consortium | 13 |
| Tong et al., 2021[61] | 1 | Yes | Recorded diagnosis of opioid use disorder (OUD) (binary) | 3379 - 256,115 | 336,711 | 5 | demographic, lab results, diagnoses, medication prescriptions | OneFlorida https://www.pcori.org/research-results/2015/oneflorida-clinical-research-consortium | 16 |
| Huang et al., 2019[58] | 2 | Yes | Mortality (binary) | 50 * 560 | 28,000 | (for each clustered community) 5, 10, 15, 35 | drug features | eICU Collaborative Research Database https://eicu-crd.mit.edu/ | 1399 |
| | | Yes | ICU stay time | 50 * 560 | 28,000 | (for each clustered community) 5, 10, 15, 35 | drug features | eICU Collaborative Research Database https://eicu-crd.mit.edu/ | 1399 |
| Duan et al., 2019[8] | 1 | Yes | Risk of fetal loss (binary) | 3557 | 35,570 | 10 | demographic, medication | No | 5 |
| Brisimi et al., 2018[46] | 2 | No | Hospitalization during a target year (binary) | NA | 45,579 | 5 | demographics, procedures, vital signs, lab test, tobacco use, medical history, past admission records | No | 215 |
| | | No | Hospitalization during a target year (binary) | NA | 45,579 | 10 | demographics, procedures, vital signs, lab test, tobacco use, medical history, past admission records | No | 215 |
| Fang et al., 2022[50] | 1 | No | NA | NA | NA | NA | NA | UCI Machine learning repository https://archive.ics.uci.edu/ml/index.php; Kaggle | NA |
| Halim et al., 2022[13] | 2 | No | Disease prediction (binary) | at least 10 per site | NA | 208 | demographics, diagnosis | eICU Collaborative Research Database https://eicu-crd.mit.edu/ | NA |
| | | No | Disease prediction (multi-label; 8) | at least 10 per site | NA | 81 | demographics, diagnosis | eICU Collaborative Research Database https://eicu-crd.mit.edu/ | NA |
| Mehta et al., 2022[52] | 2 | No | Disease (binary) | NA | 4,961 | 100 | diagnosis; questions for patients about symptoms | AI-Healthcare-Chatbot https://github.com/vsharathchandra/AI-Healthcare-chatbot | 132 |
| | | No | Disease (binary) | NA | 4,961 | 100 | diagnosis; questions for patients about symptoms | AI-Healthcare-Chatbot https://github.com/vsharathchandra/AI-Healthcare-chatbot | 132 |
| Kavitha Bharathi S et al., 2022[56] | 2 | No | Heart disease (binary) | NA | NA | 5 | demographics, vital signs, lab results, past medical/disease history) | UCI Machine learning repository https://archive.ics.uci.edu/ml/index.php | 13 |
| | | No | Heart disease (binary) | NA | NA | 5 | demographics, vital signs, lab results, past medical/disease history) | UCI Machine learning repository https://archive.ics.uci.edu/ml/index.php | 13 |
| Sadilek et al., 2021[14] | | / | Heart failure survival (binary) | / | 299 | / | demographics, vital signs, lab results, diagnosis | UCI Machine learning repository https://archive.ics.uci.edu/ml/index.php | 12 |
| | | / | Diabetes at 5-years (binary) | / | 768 | / | demographics, vital signs, lab results, past medical history | Pima Indians Diabetes Database https://www.kaggle.com/datasets/uciml/pima-indians-diabetes-database | 8 |
| | | / | SRAS-CoV-2 infection (binary) | NA | 9275 | NA | demographics, admissions, diagnosis | Malignancy in SARS-CoV2 infection https://figshare.com/articles/dataset/Malignancy_in_SARS-CoV2_infection/12666698 | 3 |
| | | / | Avian Influenza fatality infection risk (binary) | NA | 294 | NA | demographics, time from symptom to hospitalization, reported exposure | Avian influenza A(H5N1) in humans https://doi.org/10.2807/ese.16.32.19941-en | 4 |
| | | / | Relapsed infection in patients with Enterobacter bacteraemia (binary) | / | 159 | / | demographics, source of infection, co-morbid conditions, admissions, lab results, diagnosis | Risk factors for relapse or persistence of bacteraemia caused by Enterobacter https://dataverse.harvard.edu/dataset. | 12 |
| | | / | Adverse events associated with azithromycin use in infants (binary) | / | 1712 | / | demographics, recent health issues and diagnosis | MORDOR Infant Adverse Event Survey Data https://dataverse.harvard.edu/dataset.xhtml?persistentId=doi:10.7910/DVN | 0 |
| | 7 | / | Extrapulmonary tuberculosis occurrence (binary) | NA | 3342 | NA | diagnosis, demographics, site of infection, type of healthcare facility | Ghana Extra-pulmonary TB data https://dataverse.harvard.edu/file.xhtml?persistentId=doi:10.7910/DVN/TA1OII/ZSVFGO&version=1.0 | 3 |
| | | No | Hearing disability following surgery (binary) | 7,3,5 (feature dimension) | 50 | 3 | demographics, administrative, lab; administrative, demographic, | No | 15 |
| Cha et al., 2021[21] | 2 | No | Patient mortality (binary) | 3,4,9,3,3,4,6 (feature dimension) | 15,762 | 7 | administrative, demographics, lab, vital signs | eICU Collaborative Research Database https://eicu-crd.mit.edu/ | 32 |
| | | No | In hospital mortality (binary) | 1005 - 15,559 | 1,222,554 | 314 | demographics, admission type, MRCI (medication regimen complexity index score), Drugs and ICD9 codes | Premier healthcare database https://premierinc.com/newsroom?filter=category-education | 24,491 |
| Kerkouche et al., 2021[42] | 2 | No | In hospital mortality (binary) | 1005 - 15,559 | 1,222,554 | 314 | demographics, admission type, MRCI (medication regimen complexity index score), Drugs and ICD9 codes | Premier healthcare database https://premierinc.com/newsroom?filter=category-education | 24,491 |
| | | No | Clinical phenotypes | 4 mode tensor: 961x500x500x500 | | 125 | diagnosis, procedures, | CMS https://data.cms.gov/search | / |
| | | No | Clinical phenotypes | 3 mode tensor: 91999x500x500 | | 125 | diagnosis, procedures | CMS https://data.cms.gov/search | / |
| Ma et al., 2021[43] | 3 | No | Clinical phenotypes | 4 mode tensor: 272x500x500x500 | | 34 | diagnosis, procedures, | MIMIC-III | / |
| Fontana et al., 2021[16] | 1 | No | Patients' main procedure | regions based (hierarchical setting) 409, 602 - 1, 700, 000 | 4,408,710 | 11 | demographics, cause of injury, diagnosis, procedures | Texas-100 https://github.com/privacytrustlab/datasets | 85 |
| Cai et al., 2021[62] | 1 | No | Coronary artery disease status (binary) | 105 - 760 | 1267 | 4 | demographics information, lab results, medication prescriptions, ICD codes, current procedural terminology (CPT) codes; narrative features extracted via NLP | No | 533 |
| Vaid et al., 2021[39] | 2 | Yes | Mortality within seven days in hospitalized COVID-19 patients (binary) | 485 - 1644 | 4029 | 5 | demographics, past medical history, vital signs, lab results | No | 23 |
| | | Yes | Mortality within seven days in hospitalized COVID-19 patients (binary) | 485 - 1644 | 4029 | 5 | demographics, past medical history, vital signs, lab results | No | 23 |
| Lu et al., 2020[44] | 2 | No | Binary outcome (details unclear) | ~ 500 | 10,022 | 20 | general EHRs (no details) | No | 42 |
| | | No | Binary outcome (details unclear) | ~ 500 | 10,022 | 20 | general EHRs (no details) | No | 42 |
| Luo et al., 2022[25] | 1 | Yes | Cox proportional hazard (survival) | 1920 - 4078 | 14,015 | 5 | demographics, diagnosis, medical history | OneFlorida [33] | 8 |
| Choudhury et al., 2020[57] | 3 | No | Adverse drug reaction (binary) | 10,000 - 255,000 | 1,161,048 | 10 | demographics, diagnosis, prescriptions, lab results, procedures | Limited IBM MarketScan Explorys Claims-EMR Data Set https://www.ibm.com/products/explorys-ehr-data-analysis-tools | NA |
| | | No | Adverse drug reaction (binary) | 10,000 - 255,000 | 1,161,048 | 10 | demographics, diagnosis, prescriptions, lab results, procedures | Limited IBM MarketScan Explorys Claims-EMR Data Set https://www.ibm.com/products/explorys-ehr-data-analysis-tools | NA |
| | | No | Adverse drug reaction (binary) | 10,000 - 255,000 | 1,161,048 | 10 | demographics, diagnosis, prescriptions, lab results, procedures | Limited IBM MarketScan Explorys Claims-EMR Data Set https://www.ibm.com/products/explorys-ehr-data-analysis-tools | NA |
| Kim et al., 2017[19] | | No | Clinical phenotypes | NA | 38,035 | 1 ~ 5 | medications, lab results | MIMIC-III https://physionet.org/content/mimiciii | / |
| | | No | Clinical phenotypes | 3 mode tensor: 4703x748x299 | 12,725 | 2 | medications. diagnosis mode | No | / |
| | | No | Clinical phenotypes | 3 mode tensor: 8022x748x299 | 12,725 | 2 | medications. diagnosis mode | No | / |
| | | No | Clinical phenotypes | NA | 38,035 | 1 ~ 5 | medications, lab results | MIMIC-III https://physionet.org/content/mimiciii | / |
| | | No | Clinical phenotypes | 3 mode tensor: 4703x748x299 | 12,725 | 2 | medications. diagnosis mode | No | / |
| | 6 | No | Clinical phenotypes | 3 mode tensor: 8022x748x299 | 12,725 | 2 | medications. diagnosis mode | No | / |
| | | Yes | Transcatheter Aortic Value Implantation (TAVI) 1 year mortality (binary) | 631 - 1160 | 1791 | 2 | demographics, lab results, diagnoses, vital signs, medical history | No | 16 |
| | | Yes | Transcatheter Aortic Value Implantation (TAVI) 1 year mortality (binary) | 631 - 1160 | 1791 | 2 | demographics, lab results, diagnoses, vital signs, medical history | No | 16 |
| | | Yes | Transcatheter Aortic Value Implantation (TAVI) 1 year mortality (binary) | 631 - 1160 | 1791 | 2 | demographics, lab results, diagnoses, vital signs, medical history | No | 16 |
| | | Yes | Transcatheter Aortic Value Implantation (TAVI) 1 year mortality (binary) | 631 - 1160 | 1791 | 2 | demographics, lab results, diagnoses, vital signs, medical history | No | 16 |
| | | Yes | Transcatheter Aortic Value Implantation (TAVI) 1 year mortality (binary) | 631 - 1160 | 1791 | 2 | demographics, lab results, diagnoses, vital signs, medical history | No | 16 |
| Lopes et al., 2021[20] | | Yes | Transcatheter Aortic Value Implantation (TAVI) 1 year mortality (binary) | 631 - 1160 | 1791 | 2 | demographics, lab results, diagnoses, vital signs, medical history | No | 16 |
| | | Yes | Transcatheter Aortic Value Implantation (TAVI) 1 year mortality (binary) | 631 - 1160 | 1791 | 2 | demographics, lab results, diagnoses, vital signs, medical history | No | 16 |
| | 8 | Yes | Transcatheter Aortic Value Implantation (TAVI) 1 year mortality (binary) | 631 - 1160 | 1791 | 2 | demographics, lab results, diagnoses, vital signs, medical history | No | 16 |

## (b) Modeling details

| **Citation** | **No. Studies** | **Task/goal** | **Modeling approach** | **Hyperparameter methods reported** | **Performance metric(s)** |
|---|---|---|---|---|---|
| | | Association study | Quasi-Poisson regression | / | Relative bias (of log relative risk, relative to pooled analysis) |
| Edmondson et al., 2022[35] | 2 | Association study | Quasi-Poisson regression | / | Relative bias (of log relative risk, relative to pooled analysis) |
| Wang et al., 2022[1] | 1 | Prediction | Cox proportional hazards model | / | AUCROC |
| | | Model misconduct detection | Logistic regression | / | Precision, recall, F1-score (for misconduct detect correctness) |
| | | Model misconduct detection | Logistic regression | / | Precision, recall, F1-score (for misconduct detect correctness) |
| Kuo et al., 2022[15] | 3 | Model misconduct detection | Logistic regression | / | Precision, recall, F1-score (for misconduct detect correctness) |
| | | Prediction | Logistic regression | / | F-1 score, recall, precision |
| Islam et al., 2022[59] | 2 | Prediction | 3-layer multi-perceptron | No | F-1 score, recall, precision |
| | | Association Study | Poisson-Logit hurdle model | / | Relative bias (of coefficients, relative to pooled analysis) |
| Edmondson et al., 2021[54] | 2 | Association Study | Poisson-Logit hurdle model | / | Relative bias (of coefficients, relative to pooled analysis) |
| Dayan et al., 2021[33] | 1 | Prediction | Deep & Cross Network (DCN) | Yes | AUCROC |
| Kim et al., 2021[60] | 1 | Association study | Logistic regression | / | OR |
| Cui et al., 2021[30] | 1 | Prediction | Neural network | No | AUCROC, AUCPR |
| | | Prediction | Neural network | Yes | F1 score, precision, recall, and accuracy |
| Rajendran et al., 2021[17] | 2 | Prediction | Logistic regression | / | F1 score, precision, recall, and accuracy |
| Geleijnse et al., 2020[18] | 1 | Association Study | Cox proportional hazards model | / | HR |
| | | Association Study | Cox proportional hazards model | / | Relative bias (of log hazard ratios, relative to pooled analysis) |
| Duan et al., 2020[23] | 2 | Association Study | Cox proportional hazards model | / | Relative bias (of log hazard ratios, relative to pooled analysis) |
| Duan et al., 2020[34] | 1 | Association Study | Logistic regression | / | Relative bias (of coefficients, relative to pooled analysis) |
| Tong et al., 2020[63] | 1 | Association Study | Logistic regression | / | Relative bias (of coefficients, relative to pooled analysis) |
| Duan et al., 2021[24] | 1 | Association Study | Cox proportional hazards model | / | Relative bias (of coefficients, relative to pooled analysis) |
| Tong et al., 2021[61] | 1 | Association Study | Logistic regression | / | Relative bias (of coefficients, relative to pooled analysis) |
| | | Prediction | Autoencoder | Yes | AUCROC, AUCPRC |
| Huang et al., 2019[58] | 2 | Prediction | Autoencoder | Yes | AUCROC, AUCPRC |
| Duan et al., 2019[8] | 1 | Association study | Logistic regression | / | Relative difference (of ORs, relative to the pooled estimator) |
| | | Prediction | sparse SVM | Yes | AUCROC |
| Brisimi et al., 2018[46] | 2 | Prediction | sparse SVM | Yes | AUCROC |
| Fang et al., 2022[50] | 1 | Prediction | NA | / | AUCROC, AUCPRC |
| | | Prediction | HinSAGE | No | Accuracy |
| Halim et al., 2022[13] | 2 | Prediction | HinSAGE | No | Accuracy |
| Mehta et al., 2022[52] | | Prediction | Logistic regression | NA | AUCROC, Accuracy, Weighted F1 score |
| | 2 | Prediction | TabNet | No | AUCROC, Accuracy, Weighted F1 score |
| Kavitha Bharathi S et al., 2022[56] | 2 | Prediction | Logistic regression | / | Accuracy |
| | | Prediction | SVM | No | Accuracy |
| Sadilek et al., 2021[14] | 7 | Prediction | Logistic regression | / | AUCROC |
| | | Prediction | Neural network | Yes | AUCROC |
| | | Association study | Logistic regression | / | OR |
| | | Association study | Logistic regression | / | OR |
| | | Association study | Logistic regression | / | Coefficient |
| | | Association study | GLM (binomial response, log link) | / | Coefficient |
| | | Association study | Logistic regression | / | Coefficient |
| Cha et al., 2021[21] | 2 | Prediction | Autoencoder | Yes | AUCROC, Accuracy |
| | | Prediction | Autoencoder | Yes | AUCROC, Accuracy |
| Kerkouche et al., 2021[42] | 2 | Prediction | Neural Network | Yes | AUCROC, Balanced accuracy |
| | | Prediction | Neural Network | Yes | AUCROC, Balanced accuracy |
| Ma et al., 2021[43] | 3 | Phenotyping | Tensor factorization | Yes | NA |
| | | Phenotyping | Tensor factorization | Yes | NA |
| | | Phenotyping | Tensor factorization | Yes | NA |
| Fontana et al., 2021[16] | 1 | Prediction | Neural Network | Yes | F1, precision, recall |
| Cai et al., 2021[62] | 1 | Association Study | LASSO | Yes | AUCROC, Brier Score, F score (5% and 10%) |
| Vaid et al., 2021[39] | 2 | Prediction | LASSO | Yes | AUCROC |
| | | Prediction | MLP | Yes | AUCROC |
| Lu et al., 2020[44] | 2 | Prediction | Neural Network | No | NA |
| | | Prediction | Neural Network | No | NA |
| Luo et al., 2022[25] | 1 | Association Study | Cox proportional hazards model | / | Relative bias (of coefficients, relative to pooled analysis) |
| | | Prediction | SVM | No | Precision, recall, accuracy, relative error of federated and local results with respect to centralized results |
| | | Prediction | Perceptron | No | Precision, recall, accuracy, relative error of federated and local results with respect to centralized results |
| Choudhury et al., 2020[57] | 3 | Prediction | Logistic regression | / | Precision, recall, accuracy, relative error of federated and local results with respect to centralized results |
| | | Phenotyping | Tensor factorization | Yes | RMSE (between the factorized tensor and observed tensor) |
| | | Phenotyping | Tensor factorization | Yes | RMSE (between the factorized tensor and observed tensor) |
| | | Phenotyping | Tensor factorization | Yes | RMSE (between the factorized tensor and observed tensor) |
| | | Phenotyping | Tensor factorization | Yes | RMSE (between the factorized tensor and observed tensor) |
| | | Phenotyping | Tensor factorization | Yes | RMSE (between the factorized tensor and observed tensor) |
| Kim et al., 2017[19] | 6 | Phenotyping | Tensor factorization | Yes | RMSE (between the factorized tensor and observed tensor) |
| | | Prediction | Random Forest | Yes | AUCROC |
| | | Prediction | Extreme Gradient Boosting | Yes | AUCROC |
| | | Prediction | CatBoost | Yes | AUCROC |
| | | Prediction | Neural Network | Yes | AUCROC |
| | | Prediction | Random Forest | Yes | AUCROC |
| | | Prediction | Extreme Gradient Boosting | Yes | AUCROC |
| | | Prediction | CatBoost | Yes | AUCROC |
| Lopes et al., 2021[20] | 8 | Prediction | Neural Network | Yes | AUCROC |

(c) FL details

| **Citation** | **No. Studies** | **Unit** | **Countries/regions** | **Structure** | **FL framework** | **One shot** | **Topology** | **Evaluation metrics** | **Convergence Analysis (if newly developed framework and not one shot)** | **Solution for heterogeneity** | **FL vs. local** | **Code availability** |
|---|---|---|---|---|---|---|---|---|---|---|---|---|
| Edmondson et al., 2022[35] | 2 | Cross silo | United States | horizontal | ODAP (surrogate likelihood) | Yes | / | / | / | NA | No | "pda" package in R: https://github.com/Penncil/pda |
| | | Cross silo | United States | horizontal | ODAP (surrogate likelihood) | Yes | / | / | / | NA | No | "pda" package in R: https://github.com/Penncil/pda |
| Wang et al., 2022[1] | 1 | Cross silo | France, Germany, and United States | horizontal | SurvMaximin | Yes | / | NA | / | By not requiring the target population to share the same underlying model with the source population | Yes | NA |
| Kuo et al., 2022[15] | 3 | Cross silo | Artificial partitions | horizontal | GloreChain (block chain based FL) | No | Peer to peer (Decentralized) | NA | / | NA | No | NA |
| | | Cross silo | Artificial partitions | horizontal | GloreChain (block chain based FL) | No | Peer to peer (Decentralized) | NA | / | NA | No | NA |
| | | Cross silo | Artificial partitions | horizontal | GloreChain (block chain based FL) | No | Peer to peer (Decentralized) | NA | / | NA | No | NA |
| Islam et al., 2022[59] | | Cross silo | United States | horizontal | FedAvg | No | Aggregation server (Centralized) | / | / | NA | Yes | NA |
| | 2 | Cross silo | United States | horizontal | FedAvg | No | Aggregation server (Centralized) | / | / | NA | Yes | NA |
| | | Cross silo | United States | horizontal | ODAH (surrogate likelihood) | Yes | / | / | / | NA | No | "pda" package in R: https://github.com/Penncil/pda sample code for ODAH: https://PDAmethods.org/portfolio/odah/ |
| Edmondson et al., 2021[54] | 2 | Cross silo | United States | horizontal | ODAH (surrogate likelihood) | Yes | / | / | / | NA | No | "pda" package in R: https://github.com/Penncil/pda sample code for ODAH: https://PDAmethods.org/portfolio/odah/ |
| Dayan et al., 2021[33] | 1 | Cross silo | Unites States ,Canada, Taiwan, South Korea, Thailand, Brazil and UK | horizontal | FedAvg | No | Aggregation server (Centralized) | NA | / | NA | Yes | publicly available at NGC https://ngc.nvidia.com/catalog/models/nvidia:med:clara_train_covid19_exam_ehr_xray |
| Kim et al., 2021[60] | 1 | Cross silo | United States | horizontal | GLORE | No | Aggregation server (Centralized) | NA | | NA | Yes | NA |
| Cui et al., 2021[30] | 1 | Cross silo | Artificial partitions | horizontal | FeARH (anonymous random hybridization) | No | Peer-to-Peer (Decentralized) | Exchange rates | No | NA | No | NA |
| Rajendran et al., 2021[17] | | Cross silo | United States | horizontal | Single weight training | Yes | / | NA | / | NA | Yes | NA |
| | 2 | Cross silo | United States | horizontal | Cyclic weight training | No | Peer-to-Peer (Decentralized) | NA | / | NA | Yes | NA |
| Geleijnse et al., 2020[18] | 1 | Cross silo | Taiwan, Netherland | horizontal | VANTAGE6 | No | Aggregation Server (Centralized) | NA | / | NA | Yes | https://www.vantage6.ai |
| | | Cross silo | United States | horizontal | ODAC (surrogate likelihood) | Yes | / | NA | No | NA | No | "pda" package in R: https://github.com/Penncil/pda |
| Duan et al., 2020[23] | 2 | Cross silo | United States | horizontal | ODAC (surrogate likelihood) | Yes | / | NA | No | NA | No | "pda" package in R: https://github.com/Penncil/pda |
| Duan et al., 2020[34] | 1 | Cross silo | Artificial partitions | horizontal | ODAL2 (surrogate likelihood) | Yes | / | / | No | NA | Yes | https://pdamethods.org/ |
| Tong et al., 2020[63] | 1 | Cross silo | United States, Japan | horizontal | Robust-ODAL (surrogate likelihood) | Yes | / | / | No | Using median instead of mean of the intermediates values to make algorithm more robust to heterogeneity | No | NA |
| Duan et al., 2021[24] | 1 | Cross silo | United States | horizontal | ODAC (surrogate likelihood) | Yes | / | / | No | NA | No | NA |
| Tong et al., 2021[61] | 1 | Cross silo | United States | horizontal | ODAL2 (surrogate likelihood) | Yes | / | / | No | NA | No | NA |
| | | Cross silo | United States | horizontal | Community based FL | No | Aggregation server (Centralized) | Communication rounds | / | NA | No | NA |
| Huang et al., 2019[58] | 2 | Cross silo | United States | horizontal | Community based FL | No | Aggregation server (Centralized) | Communication rounds | / | NA | No | NA |
| Duan et al., 2019[8] | 1 | Cross silo | United States | horizontal | ODAL (surrogate likelihood) | Yes | / | / | No | NA | No | https://rdrr.io/github/Penncil/pda/src/R/ODAL.R |

| | | | | | | | | | | | | | |
|---|---|---|---|---|---|---|---|---|---|---|---|---|---|
| Brisimi et al., 2018[46] | 2 | Cross silo | Artificial partitions | horizontal | Cluster Primal Dual Splitting (cPDS) | No | Peer-to-Peer (Decentralized) | Number of iterations, computation cost, communication cost | Yes | NA | No | NA |
| | | Cross silo | Artificial partitions | horizontal | Cluster Primal Dual Splitting (cPDS) | No | Peer-to-Peer (Decentralized) | Number of iterations, computation cost, communication cost | Yes | NA | No | NA |
| Fang et al., 2022[50] | 1 | NA | Artificial partitions | horizontal | FDP-CTGAN | No | Aggregation server (Centralized) | NA | No | NA | No | https://github.com/juliecious/CTGAN/tree/DP |
| Halim et al., 2022[13] | 2 | Cross silo | Artificial partitions | horizontal | FedAvg + Multi-KRUM | No | Aggregation server (Centralized) | NA | / | By introducing extra weight matrices | No | / |
| | | Cross silo | Artificial partitions | horizontal | FedAvg + Multi-KRUM | No | Aggregation server (Centralized) | NA | / | By introducing extra weight matrices | No | / |
| Mehta et al., 2022[52] | 2 | Cross silo | Artificial partitions | horizontal | FedProx | No | Aggregation server (Centralized) | NA | / | NA | No | NA |
| | | Cross silo | Artificial partitions | horizontal | FedProx | No | Aggregation server (Centralized) | NA | / | NA | No | NA |
| Kavitha Bharathi S et al., 2022[56] | 2 | Cross silo | Artificial partitions | horizontal | FedAvg | No | Aggregation server (Centralized) | Number of iterations | / | NA | No | NA |
| | | Cross silo | Artificial partitions | horizontal | FedAvg | No | Aggregation server (Centralized) | Number of iterations | / | NA | No | NA |
| Sadilek et al., 2021[14] | | Cross patient | Per patient | horizontal | FedAvg | No | Aggregation server (Centralized) | Runtime | Yes | NA | Yes | https://github.com/google/federated-ml-health |
| | | Cross patient | Per patient | horizontal | FedAvg | No | Aggregation server (Centralized) | Runtime | Yes | NA | Yes | https://github.com/google/federated-ml-health |
| | | Both | Per patient; artificial partitions based on groups | horizontal | FedAvg | No | Aggregation server (Centralized) | Runtime | Yes | NA | Yes | https://github.com/google/federated-ml-health |
| | | Both | Per patient; artificial partitions based on groups | horizontal | FedAvg | No | Aggregation server (Centralized) | Runtime | Yes | NA | Yes | https://github.com/google/federated-ml-health |
| | | Cross patient | Per patient | horizontal | FedAvg | No | Aggregation server (Centralized) | Runtime | Yes | NA | Yes | https://github.com/google/federated-ml-health |
| | | Cross patient | Per patient | horizontal | FedAvg | No | Aggregation server (Centralized) | Runtime | Yes | NA | Yes | https://github.com/google/federated-ml-health |
| | 7 | Both | Per patient; artificial partitions based on groups | horizontal | FedAvg | No | Aggregation server (Centralized) | Runtime | Yes | NA | Yes | https://github.com/google/federated-ml-health |
| | | Cross silo | Artificial partitions | vertical | NA | NA | NA | NA | No | NA | / | Available from the authors upon request |
| Cha et al., 2021[21] | 2 | Cross silo | Artificial partitions | vertical | NA | NA | NA | NA | No | NA | / | Available from the authors upon request |
| Kerkouche et al., 2021[42] | | Cross silo | United states | horizontal | FL-SIGN | No | Aggregation server (Centralized) | Average bandwidth consumption | Yes | Downsampling (for imbalance data) | No | https://github.com/raouf-kerkouche/Privacy-preserving-and-Bandwith-Efficient-Federated-Learning-An-Application-to-In-Hospital-Mortality |
| | 2 | Cross silo | United states | horizontal | FL-SIGN-DP | No | Aggregation server (Centralized) | Average bandwidth consumption | Yes | Downsampling (for imbalance data) | No | https://github.com/raouf-kerkouche/Privacy-preserving-and-Bandwith-Efficient-Federated-Learning-An-Application-to-In-Hospital- |
| | | Cross silo | Artificial partitions | horizontal | 1) FedGTF-EF: Communication Efficient GTF with Block Randomization, Gradient Compression and Error-Feedback 2) FedGTF-EF-PC: Further Reducing Communication Cost by Periodic Communication | No | Aggregation Server (Centralized) | Computation time, Communication Cost | Yes | NA | No | NA |
| | | Cross silo | Artificial partitions | horizontal | 1) FedGTF-EF: Communication Efficient GTF with Block Randomization, Gradient Compression and Error-Feedback 2) FedGTF-EF-PC: Further Reducing Communication Cost by Periodic Communication | No | Aggregation Server (Centralized) | Computation time, Communication Cost | Yes | NA | No | NA |
| Ma et al., 2021[43] | | Cross silo | Artificial partitions | horizontal | 1) FedGTF-EF: Communication Efficient GTF with Block Randomization, Gradient | No | Aggregation Server (Centralized) | Computation time, Communication Cost | Yes | NA | No | NA |
| | 3 | | | | Compression and Error-Feedback 2) FedGTF-EF-PC: Further Reducing Communication Cost by Periodic Communication | | | | | | | |
| Fontana et al., 2021[16] | 1 | Cross silo | United States | horizontal | HOLDA (Hierarchical crOss-siLo feDerated Averaging) | No | Aggregation Server (Centralized) | NA | No | NA | Yes | NA |
| Cai et al., 2021[62] | 1 | Cross silo | NA | horizontal | Data-Shielding High-dimensional Integrative Regression (SHIR) | Yes | / | / | / | Meta-analysis debiasing (for cross-study heterogeneous effects) | Yes | NA |
| | | Cross silo | United States | horizontal | FedAvg | No | Aggregation Server (Centralized) | NA | / | NA | Yes | NA |
| Vaid et al., 2021[39] | 2 | Cross silo | United States | horizontal | FedAvg | No | Aggregation Server (Centralized) | NA | / | NA | Yes | NA |
| | | Cross silo | NA | horizontal | Fully decentralized non-convex stochastic gradient descent | No | Peer-to-Peer (Decentralized) | Optimality gap, Number of iterations, Communic ation rounds | Yes | NA | No | NA |
| Lu et al., 2020[44] | 2 | Cross silo | NA | horizontal | Fully decentralized non-convex stochastic gradient tracking | No | Peer-to-Peer (Decentralized) | Optimality gap, Number of iterations, Communic ation rounds | Yes | Stochastic Gradient Tracking | No | NA |
| Luo et al., 2022[25] | 1 | Cross silo | United States | horizontal | ODACH (surrogate likelihood) | Yes | / | / | / | Sharing second-order gradient of surrogate likelihood (for heterogeneous baseline hazard functions and covariate distributions) | No | NA |
| | | Cross silo | Artificial partitions | horizontal | FedAvg (ClassRatio & Loss per sample) | No | Aggregation Server (Centralized) | NA | No | Assign weights based on class ratio; Consider loss per sample | Yes | NA |
| | | Cross silo | Artificial partitions | horizontal | FedAvg (ClassRatio & Loss per sample) | No | Aggregation Server (Centralized) | NA | No | Assign weights based on class ratio; Consider loss per | Yes | NA |
| Choudhury et al., 2020[57] | 3 | Cross silo | Artificial partitions | horizontal | FedAvg (ClassRatio & Loss per sample) | No | Aggregation Server (Centralized) | NA | No | Assign weights based on class ratio; Consider loss per sample | Yes | NA |
| | | Cross silo | Artificial partitions | horizontal | Privacy Preserving Computational Phenotyping | No | Aggregation Server (Centralized) | Computatio n time, Communic | Yes | NA | Yes | NA |
| | | Cross silo | Artificial partitions | horizontal | Privacy Preserving Computational Phenotyping | No | Aggregation Server (Centralized) | Computatio n time, Communic | Yes | NA | Yes | NA |
| | | Cross silo | Artificial partitions | horizontal | Privacy Preserving Computational Phenotyping | No | Aggregation Server (Centralized) | Computatio n time, Communic | Yes | NA | Yes | NA |
| | | Cross silo | United States | horizontal | Privacy Preserving Computational Phenotyping | No | Aggregation Server (Centralized) | Computatio n time, Communic | Yes | NA | Yes | NA |
| | | Cross silo | United States | horizontal | Privacy Preserving Computational Phenotyping | No | Aggregation Server (Centralized) | Computatio n time, Communic | Yes | NA | Yes | NA |
| Kim et al., 2017[19] | 6 | Cross silo | United States | horizontal | Privacy Preserving Computational Phenotyping | No | Aggregation Server (Centralized) | Computatio n time, Communic | Yes | NA | Yes | NA |
| | | Cross silo | Netherlands | horizontal | Cyclic weight transfer | No | Peer-to-Peer (Decentralized) | NA | / | Class weighting & random over sampling (for imbalance data) | Yes | NA |
| | | Cross silo | Netherlands | horizontal | Cyclic weight transfer | No | Peer-to-Peer (Decentralized) | NA | / | Class weighting & random over sampling (for imbalance data) | Yes | NA |
| | | Cross silo | Netherlands | horizontal | Cyclic weight transfer | No | Peer-to-Peer (Decentralized) | NA | / | Class weighting & random over sampling (for imbalance data) | Yes | NA |
| | | Cross silo | Netherlands | horizontal | Cyclic weight transfer | No | Peer-to-Peer (Decentralized) | NA | / | Class weighting & random over sampling (for imbalance data) | Yes | NA |
| | | Cross silo | Netherlands | horizontal | Stacking | Yes | / | / | / | Class weighting & random over sampling (for imbalance data) | Yes | NA |
| | | Cross silo | Netherlands | horizontal | Stacking | Yes | / | / | / | Class weighting & random over sampling (for imbalance data) | Yes | NA |
| | | Cross silo | Netherlands | horizontal | Stacking | Yes | / | / | / | Class weighting & random over sampling (for imbalance data) | Yes | NA |
| Lopes et al., 2021[20] | 8 | Cross silo | Netherlands | horizontal | Stacking | Yes | / | / | / | Class weighting & random over sampling (for imbalance data) | Yes | NA |